\documentclass{article}

\PassOptionsToPackage{table}{xcolor}
\usepackage{etoolbox}       %
\usepackage[utf8]{inputenc} %

\usepackage{graphicx}  %

\usepackage{amsmath,amsfonts,bm}

\newcommand{\m}{\mathbf{m}}

\def\eqref#1{equation~(\ref{#1})}

\def\1{\bm{1}}

\DeclareMathAlphabet{\mathsfit}{\encodingdefault}{\sfdefault}{m}{sl}
\SetMathAlphabet{\mathsfit}{bold}{\encodingdefault}{\sfdefault}{bx}{n}

\newcommand{\R}{\mathbb{R}}

 \providecommand{\ttheta}{\bm{\theta}}

  \providecommand{\cL}{\mathcal{L}}

\newcommand{\mahdi}[1]{}

\def\remark{\addtocounter{remark}{1}\def\@currentlabel{\theremark}%
\emph{Remark~\theremark}. } \makeatother
\newcounter{remark}

 \usepackage{color-edits}
 \addauthor{mh}{red}
 \addauthor{xw}{magenta}
 \addauthor{ne}{brown}
 
\usepackage{anyfontsize}

\usepackage[parfill]{parskip}
\usepackage{amsmath,amsthm}
\usepackage{mathtools}  %
\usepackage{adjustbox}
\usepackage{threeparttable}
\usepackage{tablefootnote}
\usepackage{afterpage}

  \usepackage[parfill]{parskip}
  \usepackage[citestyle=authoryear,sorting=nyt,maxbibnames=99, maxcitenames=1, backend=biber, uniquename=false]{biblatex}
  \newcommand{\citep}{\parencite}
  
  \newcommand{\citet}{\textcite}
  \newlength{\defbaselineskip}
  \RequirePackage[T1]{fontenc}
  \RequirePackage[tt=false, type1=true]{libertine}
  \RequirePackage[varqu]{zi4}
  \RequirePackage[libertine]{newtxmath}

\usepackage{bm}

\usepackage[inline]{enumitem}
\usepackage{booktabs}
\usepackage{subcaption}
\usepackage{multirow}
\usepackage{wrapfig}
\usepackage{algorithm,algorithmicx,algpseudocode}
\usepackage{nicematrix}

\usepackage[frozencache,cachedir=.]{minted} %

\usepackage{import}
\usepackage{lettrine}

\usepackage{pifont}%

\usepackage[T1]{fontenc}

\usepackage{thm-restate}

\usepackage{url}            %
\usepackage{booktabs}       %
\usepackage{amsfonts}       %
\usepackage{nicefrac}       %
\usepackage{microtype}      %

\usepackage{wrapfig}

\usepackage{caption}
\usepackage{capt-of}
\usepackage{lipsum}

\usepackage[colorlinks]{hyperref}
\usepackage[capitalise,noabbrev]{cleveref}

\usepackage{threeparttable}

\definecolor{c1}{HTML}{586770}
\definecolor{c4}{HTML}{2a4a67}
\definecolor{c3}{HTML}{6d2a58}
\definecolor{c2}{HTML}{34142a}
\hypersetup{colorlinks=true, citecolor=c3, linkcolor=c2, urlcolor=c2}

\definecolor{myblue}{HTML}{FDF5E0} 
\definecolor{mygray}{HTML}{DBE2E9} 
\definecolor{mygreen}{HTML}{E6F3FC}

\definecolor{dark2orange}{rgb}{0.9, 0.4, 0.}
\definecolor{dark2purple}{rgb}{0.4, 0.4, 0.8}

\hypersetup{
    colorlinks=true,
    linkcolor=black,
    urlcolor=c1,
    citecolor=c1,
}

\usepackage{multirow}

\newcommand{\M}[0]{\mathcal{M}}

\usepackage{mathtools}

\usepackage{tikz}

\usepackage{authblk}
\usepackage{epigraph}
\usepackage{soul}

\usepackage{tcolorbox}
\usepackage{xcolor}
\usepackage{lettrine,Zallman}
\newtcolorbox{c4box}{boxrule=1pt, colback=c4!5!white,colframe=c4!50!white}
\newtcolorbox{c3box}{boxrule=1pt, colback=c3!5!white,colframe=c3!50!white}
\newtcolorbox{c2box}{boxrule=1pt, colback=c2!5!white,colframe=c2!50!white}
\newtcolorbox{c1box}{boxrule=1pt, colback=c1!5!white,colframe=c1!50!white}

\newcommand\blfootnote[1]{%
  \begingroup
  \renewcommand\thefootnote{}\footnote{#1}%
  \addtocounter{footnote}{-1}%
  \endgroup
}

\title{TNT: Improving Chunkwise Training for Test-Time Memorization}

\author{%
Zeman Li$^{1,2}$\thanks{Work done while interning at Google Research.} \quad Ali Behrouz$^{2}$ \quad Yuan Deng$^{2}$ \quad Peilin Zhong$^2$ \quad Praneeth Kacham$^{2}$ \quad Mahdi Karami$^{2}$ \quad Meisam Razaviyayn$^{1,2}$ \quad Vahab Mirrokni$^2$ \\
$^1$University of Southern California \quad $^2$Google Research\\
\protect \blfootnote{\texttt{\{zemanli, alibehrouz, dengyuan, pkacham, mahdika, razaviyayn, mirrokni\}@google.com}, and peilin.zhong@columbia.edu}
}
\date{}

\begin{document}

\vspace{-20ex}
\maketitle

\begin{abstract}

Recurrent neural networks (RNNs) with deep test-time memorization modules, such as Titans and TTT, represent a promising, linearly-scaling paradigm distinct from Transformers. While these expressive models do not yet match the peak performance of state-of-the-art Transformers, their potential has been largely untapped due to prohibitively slow training and low hardware utilization.
Existing parallelization methods force a fundamental conflict governed by the chunksize hyperparameter: large chunks boost speed but degrade performance, necessitating a fixed, suboptimal compromise. To solve this challenge, we introduce TNT, a novel training paradigm that decouples training efficiency from inference performance through a two-stage process. Stage one is an efficiency-focused pre-training phase utilizing a hierarchical memory. A global module processes large, hardware-friendly chunks for long-range context, while multiple parallel local modules handle fine-grained details. Crucially, by periodically resetting local memory states, we break sequential dependencies to enable massive context parallelization. Stage two is a brief fine-tuning phase where only the local memory modules are adapted to a smaller, high-resolution chunksize, maximizing accuracy with minimal overhead. Evaluated on Titans and TTT models, TNT achieves a substantial acceleration in training speed—up to 17$\times$ faster than the most accurate baseline configuration—while simultaneously improving model accuracy. This improvement removes a critical scalability barrier, establishing a practical foundation for developing expressive RNNs and facilitating future work to close the performance gap with Transformers.

\end{abstract}

\section{Introduction}
\label{sec:intro}
The demand for modeling long sequences highlights a fundamental limitation of standard softmax attention ~\citep{vaswani2017attention}: its quadratic complexity bottlenecks scaling. This has spurred extensive research into more efficient architectures.

Among these emerging paradigms, a particularly powerful approach is rooted in test-time memorization~\citep{sun2024learning}. Architectures leveraging this principle, which we refer to as \textbf{deep memory modules}, utilize a deep, online-adapted sub-network whose parameters are rapidly updated to encode in-context information. Prominent examples include Titans~\citep{behrouz2024titans} and Atlas~\citep{behrouz2025atlas}. This method stands in sharp contrast to \textbf{linear memory modules}~\citep{yang2024gated,yang2024parallelizing,dao2024transformers,sun2023retentive}, which, despite their efficiency, are constrained by matrix-valued hidden states and linear state transitions. By leveraging expressive non-linear objectives and update rules, deep memory modules can theoretically overcome these limitations. While these methods generally do not yet achieve the state-of-the-art performance of Transformers, they represent a potentially promising paradigm for efficient sequence modeling, provided their training bottlenecks can be resolved.

\begin{figure}[!t]
        \centering
        \includegraphics[width=1\linewidth]{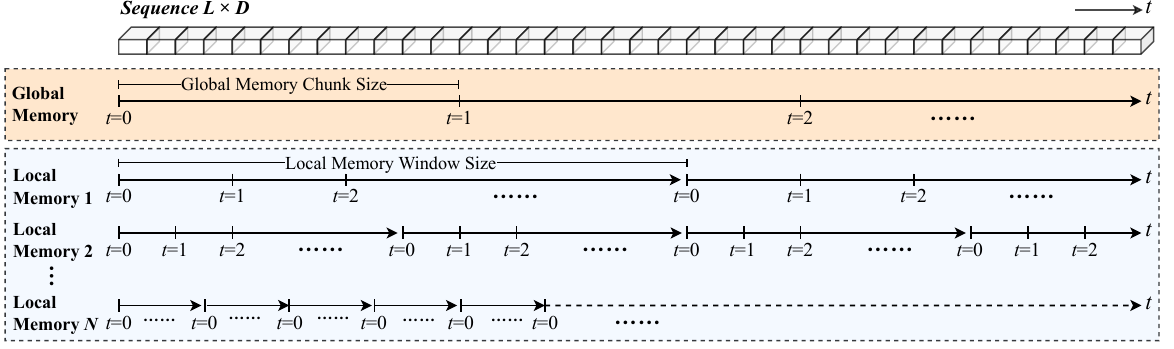}
        \caption{The basic diagram for illustraing TNT memory hierarchy. In each row, the updates at the same value of $t$ ran at the same time (run in parrallel). $t=0$ is the initialization of the memory.
        }
        \label{fig:tnt_stage_1_hierarchy}
\end{figure}

Despite their expressive advantages, deep memory modules lack the efficient training algorithms of their linear counterparts, leading to low hardware utilization. Unlike linear memory modules, which utilize hardware-efficient parallelization, deep memory modules face challenges stemming from non-linear recurrences (e.g., LayerNorm between chunks) and the complexity of their deep structures.
In practice, these challenges constrain their training to more frequent online updates on small data segments, resulting in poor computational throughput in training. This creates an inherent tension, as these models typically rely on a fixed, small chunk size (e.g., 16 to 64 tokens) to balance memory layer expressiveness against training efficiency. Consequently, this trade-off between in-context learning capability and computational performance has become a critical bottleneck preventing the application of these models to truly long sequences in practice. Resolving this fundamental tension is the primary goal of this work.

Recent work attempts to mitigate this issue. \citet{zhang2025test} combines large chunks with local attention to enhance parallelism. However, this circumvents the inefficiency rather than solving it, complicates the analysis by mixing memory and attention, and neglects the need for small chunks (ideally 1) during inference. Concurrently, \citet{guo2025log}  proposed a hierarchical memory system, but it is limited to linear memory modules and does not support short-term memories.

To resolve this tension, we introduce TNT\footnote{TNT can be viewed as an abbreviation of \textit{Titans iNside Titans} or \textit{TTT iNside TTT}. It also hints to its ``explosive" impact on training efficiency.}, a novel training paradigm for deep memory modules. Our core insight is that different components of the model should process information at different granularities during distinct training stages. TNT begins with an \textbf{efficiency-focused pre-training stage} designed to maximize throughput. This is achieved via a hierarchical memory system: a \textbf{global memory} module operates on large, hardware-friendly chunks to capture long-range context, 
while multiple \textbf{local memory} modules handle fine-grained details in parallel. 
 Crucially, we introduce a periodic reset mechanism for the local memory states. This breaks the sequential dependencies inherent even in non-linear RNNs (e.g., those with normalization between steps), enabling massive context parallelization. This is a key innovation, as efficiently parallelizing \textit{non-linear} recurrences across the sequence length is a long-standing challenge, largely unsolved outside of Transformers and specialized linear RNNs (where parallel scans apply).
Subsequently, a \textbf{performance-focused fine-tuning stage} adapts the model for optimal inference. During this stage, only the local memory modules are adjusted to use smaller chunk sizes, achieving high-resolution accuracy with minimal additional computational cost. This two-stage approach effectively decouples training efficiency from inference performance, significantly improving training scalability while addressing a key limitation of prior architectures. Furthermore, the local memory system itself can be hierarchical, employing multiple modules operating at different resolutions. This \textit{multi-resolution} approach allows the model to capture \textit{complex, multi-scale temporal dynamics} more effectively than a single fixed chunk size.

TNT is a general training paradigm applicable to any deep memory module rather than a specific architecture. By decoupling training throughput from inference accuracy, we resolve a fundamental tension constraining prior work. This removes dependency on hardware-specific optimizations for small chunks and enables flexible exploration of the architectural design space. We believe this paradigm will open new research avenues towards replacing softmax attention.

\begin{wrapfigure}{r}{0.58\textwidth} 
    \centering
    \includegraphics[width=1\linewidth]{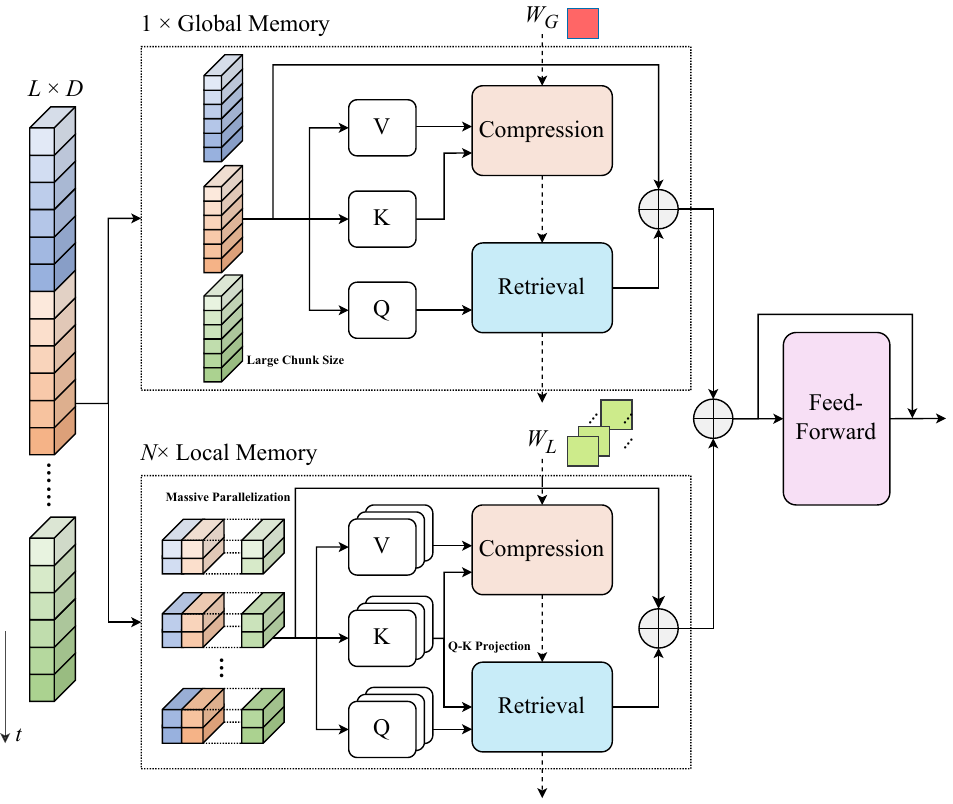}
    \vspace{-0.7cm}
    \caption{Architectural overview of TNT Stage 1.} %
    \label{fig:tnt_stage_1}
    \vspace{-0.2cm}
\end{wrapfigure}

Our main contributions are summarized as follows:
\begin{itemize} [leftmargin=*]
    \item We identify three fundamental challenges limiting the scalability and performance of deep memory modules: 1) domain mismatch between memory compression and retrieval; 2) tradeoff between memory performance and computational efficiency; 3)  chunksize mismatch between training and inference (Section~\ref{sec: challenges}).
    \item We introduce Q-K Projection, an efficient mechanism to resolve the domain mismatch between memory compression and retrieval (Section~\ref{subsubsec:MemRetrieval}).
    \item We introduce a novel hierarchical memory architecture with periodic state resets, enabling context parallelism for non-linear deep memory modules (Section~\ref{subsec:Stage1}). 
   \item We introduce an efficient fine-tuning mechanism to address chunksize mismatch between training and inference in deep memory modules (Section~\ref{subsec:Stage2}).
    \item Putting all above together, we introduce TNT, a general two-stage training paradigm that decouples training efficiency from inference performance by combining efficient pre-training with high-resolution fine-tuning (Figure~\ref{fig:tnt_stage_1_hierarchy}, Figure~\ref{fig:tnt_stage_1}, Section~\ref{sec: TNT}).
    \item   We validate TNT on the Titans architecture, achieving up to a $17.37\times$ training speedup while improving accuracy, significantly advancing the practicality of expressive RNNs (Section~\ref{sec:experiments}).
\end{itemize}

\paragraph{Problem Definition and Notations}
We aim to train a neural network with parameters $\ttheta \in \R^{d_m}$ to perform next-token prediction. Given a sequence $\mathbf{x} = (x_1, \ldots, x_L)$, the model's objective is to predict each token $x_t$ using the context of its preceding tokens $(x_1, \ldots, x_{t-1})$. Following the attention formulation, each token $x_t$ is represented by a $d$-dimensional vector. Each input token $\mathbf{x}_t$ is projected into query, key, and value vectors: $q_t, k_t, v_t \in \mathbb{R}^d$.  For ease of notation in subsequent chunkwise operations, we define a function $\xi(i, j) := i - (i \mod j)$, which finds the beginning of the chunk containing index $i$ for a chunk size $j$.

\section{Preliminary} \label{sec: preliminary}
This section reviews preliminaries. Expanded  related work is in \autoref{app:rw}.

\subsection{Deep Memory Modules via Test-Time Memorization} \label{sec: deep-memory-modules}

A powerful paradigm for sequence modeling is Test-Time Memorization~\citep{sun2024learning}, which enhances models by incorporating a secondary, rapidly adaptable neural network. Unlike the primary model parameters, or ``slow weights'' ($\theta$) updated only during training, this approach introduces ``fast weights''~\citep{schlag2021linear}. These fast weights, denoted by $W$, parameterize a sub-network, $f(W, \cdot): \mathbb{R}^d \rightarrow \mathbb{R}^d$, that is updated online-during both training and inference-based on incoming tokens to dynamically store contextual information. 
While these modules do not yet achieve SOTA results compared to Transformers~\citep{arora2024simple, behrouz2025atlas}, improving their training efficiency is crucial for enabling the wider experimentation needed to close this gap.

In this work, we focus on a similar/relevant principle: \textbf{deep memory modules}~\citep{irie2021going, sun2024learning, behrouz2024titans, behrouz2025atlas, behrouz2025s}. In contrast to \textbf{linear memory modules}~\citep{sun2023retentive, yang2024parallelizing,dao2024transformers, karami2025lattice, hu2025comba}, which are characterized by linear state transitions, deep memory modules employ non-linear recurrence rules and complex memory structures.

The core mechanism of a deep memory module can be distilled into two sequential operations for each input token: \emph{1. Memory Compression} and \emph{2. Memory Retrieval}. These are formally defined as:
\begin{align}
    \emph{Memory Compression:} \quad & W_t \leftarrow W_{t-1} - \eta_t \nabla_W \mathcal{L}\big(f(W_{t-1}, k_t), v_t\big) %
    \label{eq:ttt_memory_compress} \\
    \emph{Memory Retrieval:} \quad & o_t = f(W_{t}, q_t) %
    \label{eq:ttt_memory_retrieval}
\end{align}

 In \emph{Memory Compression}, the fast weights $W$ are updated via gradient descent, guided by a self-supervised loss $\mathcal{L}(\cdot,\cdot)$ (e.g., MSE) and a learned learning rate $\eta_{t}$. The objective associates a transformed key, $f(W_{t-1},k_{t})$, with its value, $v_{t}$, compressing information into the fixed-size neural memory~\citep{wang2025testtimeregressionunifyingframework, behrouz2025itsconnectedjourneytesttime}. In \emph{Memory Retrieval}, the updated $W_{t}$ processes a query $q_{t}$ to produce $o_{t}$. These two operations are performed iteratively for each token.

\subsection{Chunkwise Parallel Training} \label{sec: chunkwise-parrallel training}

The sequential dependency ($W_{t}$ depends on $W_{t-1}$) in Eqs. \ref{eq:ttt_memory_compress}-\ref{eq:ttt_memory_retrieval} prevents parallelization across the sequence length. To address this, deep memory modules adopt chunkwise parallel training ~\citep{hua2022transformer, sun2023retentive} to enable hardware-efficient training.

The core principle is to divide the input sequence into non-overlapping chunks of size $C$. Within each chunk, an approximation of the gradient of the loss for every token is computed with respect to the fast-weight state from the beginning of that chunk. This formulation breaks the sequential token-to-token dependency for gradient calculation, which allows the updates for all tokens within the chunk to be computed in parallel. The formal operations for a token at time step $t$ are as follows:
\vspace{-0.3cm}
\begin{align}
    \emph{Chunkwise Memory Compression:} \quad & W_t \leftarrow W_{\xi(t,C)} - \sum^{t}_{\tau=\xi(t,C)}\eta_\tau \nabla_{W} \mathcal{L}\big(f(W_{\xi(t,C)},k_\tau), v_\tau \big) \label{eq:chunkwise_memory_compress} \\
    \emph{Chunkwise Memory Retrieval:} \quad & o_t = f(W_{t}, q_t) \label{eq:chunkwise_memory_retrieval}
\end{align}
Here, $W_{\xi(t,C)}$ denotes the state of the fast weights at the start of the chunk containing token $t$ (See the definition of $\xi(\cdot,\cdot)$  at the end of Section~\ref{sec:intro}). 
Although the update to obtain $W_t$ still depends on prior tokens within its chunk, the summation of gradients can be implemented efficiently using parallel operations (e.g., cumulative summation), significantly improving hardware utilization during training. However, a sequential dependency remains: the final state of the fast weights from the $n$-th chunk, $W_{nC}$, is used as the initial state for the $(n+1)$-th chunk.

\section{Challenges in Deep Memory Modules} \label{sec: challenges}
While chunkwise parallelization enables deep memory modules to train on long sequences, this paradigm introduces significant challenges that limit their practical performance and scalability. In this section, we outline three fundamental challenges with deep memory modules.

\paragraph{Challenge 1: Lack of Efficient Training Implementations.}
A primary challenge for deep memory modules is the inefficiency of their training process, which leads to poor hardware utilization. While chunkwise parallelization theoretically enables sub-quadratic scaling, in practice, the training throughput lags significantly behind that of linear memory modules. This discrepancy arises from a fundamental tension between model expressiveness and computational efficiency.
\begin{wrapfigure}{r}{0.5\textwidth}
    \centering
    \vspace{-0.15cm}
    \includegraphics[width=0.5\textwidth]{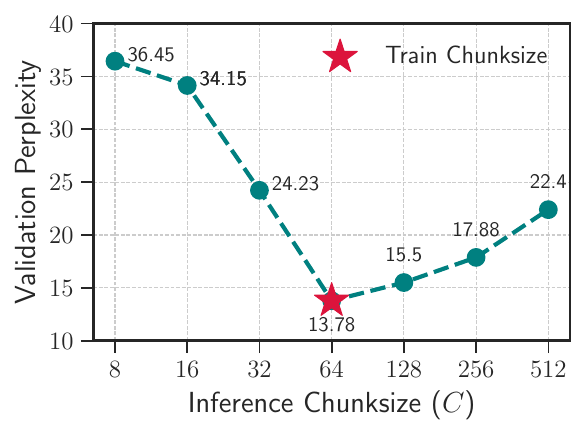}
    \vspace{-0.8cm}
    \caption{Sensitivity of inference chunk size on a $550$M Titans model pre-trained with $C=64$. Performance is optimal when the inference chunk size matches the training one.}
    \label{fig:motivation_stage_2}
    \vspace{-0.3cm}
\end{wrapfigure}

To maintain a fine-grained learning signal, deep memory modules require small chunk sizes (e.g., 16-64 tokens) ~\citep{sun2024learning}, which fail to saturate accelerators, making training memory-bound (rather than compute-bound). While linear memory modules use customized kernels (e.g., leveraging SRAM) ~\citep{sun2023retentive, gu2023mamba,qinvarious,yang2024gated,yangparallelizing}, this relies on linear state transitions and is incompatible with the large, non-linear states of deep memory modules.

The consequence is that deep memory modules suffer from extremely low FLOPs utilization, often falling below 5-10\% of peak hardware performance~\citep{zhang2025test}. This severe inefficiency makes pre-training prohibitively slow and costly, creating a major bottleneck that undermines the practical application of these expressive models to truly long sequences.

\paragraph{Challenge 2: Inconsistency Between Memory Compression and Retrieval.}
A fundamental inconsistency exists between how the memory sub-network is trained and how it is utilized. During Memory Compression (Eq.~\ref{eq:ttt_memory_compress}), the sub-network $f(W, \cdot)$ is optimized to learn a mapping from the key space to the value space by associating keys ($k_t$) with values ($v_t$). However, during Memory Retrieval (Eq.~\ref{eq:ttt_memory_retrieval}), the network is queried using a query vector ($q_t$) instead of a key. This substitution violates the intended input domain of the learned function, creating a discrepancy between the training objective and the retrieval task. This domain shift can degrade the integrity of the learned mapping and limit the model's retrieval performance. Our empirical validation can be found in Section~\ref{sec: ablation_study}

\paragraph{Challenge 3: Performance Sensitivity to a Fixed Pre-training Chunksize.}
The chunk size hyperparameter, $C$, governs the trade-off between training throughput and model expressiveness. Current practice for deep memory modules is to use the same fixed chunk size for both pre-training and inference. However, we observe that inference-time performance is highly sensitive to this pre-training choice. For example, as shown in Figure~\ref{fig:motivation_stage_2}, a model pre-trained with a chunk size of 64 achieves optimal perplexity only when evaluated with that same chunk size.

This result reveals a critical train-test mismatch and contradicts the intuition that smaller chunks at inference should yield superior performance by capturing fine-grained dependencies with a "fresher" learning signal. Instead, the model becomes over-specialized to the specific chunk resolution seen during training. This inflexibility is a significant limitation; ideally, a model pre-trained with a large, hardwarefriendly chunk size should be adaptable enough to perform even better with smaller, more precise chunk sizes at inference. Current deep memory modules fail to achieve this adaptability.

\section{TNT: An Improved Training Framework for Deep Memory} \label{sec: TNT}

To address the challenges outlined in Section~\ref{sec: challenges}, we introduce TNT, an improved training paradigm for deep memory modules. Our framework is structured around a two-stage process designed to resolve the inherent tension between training efficiency and inference performance: an \textbf{Efficiency-focused Pre-training Stage} and a \textbf{Performance-focused Fine-tuning Stage}.

The first stage maximizes training throughput by introducing a novel hierarchical memory architecture that enables unprecedented parallelism, directly addressing the challenges of low hardware utilization and inconsistent memory objectives (Challenges 1-2). The second stage employs an efficient fine-tuning strategy that adapts the model to high-resolution, small-chunk inference, resolving the sensitivity to the pre-training chunk size (Challenge 3). This two-stage approach effectively decouples training efficiency from inference performance, overcoming a key limitation of prior deep memory architectures.

\subsection{TNT Stage 1: Efficient-focused Pre-training}
\label{subsec:Stage1}
\subsubsection{TNT Memory Compression: Hierarchy Memory}\label{subsubsec:MemCompressionl}

Sequential state dependency prevents context parallelism (processing sequence shards in parallel across devices). To enable this, we propose that all parallel shards initialize their \textbf{local memory} with the same learned state, $W_{init}$. This breaks inter-chunk dependency, allowing massive parallelization. However, this causes local memory modules to lose the global context.
To solve this, we introduce a \textbf{global memory} module, parameterized by $V$, that operates in parallel with the sharded local memories. The global memory processes the sequence with a relatively large chunk size (e.g., 2048 or greater), allowing it to efficiently capture long-range dependencies while maintaining high hardware utilization. This creates a hierarchical system where local memories handle fine-grained information within parallel shards, while the global memory provides the overarching context.

This hierarchical structure is flexible; a model can be designed with 1 global and $N$ local memory modules, each operating at a different resolution. For clarity of illustration, we will assume the simplest case where $N=1$. We defer the generalized formulation of TNT to Appendix~\ref{app:tnt_complete}. We now formally define our memory compression mechanism.

\begin{c2box}
\noindent
\textbf{TNT Memory Compression Rule.} The hierarchical memory is updated as follows:

\paragraph{Global Memory Update.} The global memory state $V$ evolves sequentially across the input with a large chunk size $C_G$. 
\begin{align} \label{eq:tnt_global_memory_compress}
V_{(n+1)C_G}  \gets V_{nC_G} - \sum_{t=kC_G}^{(k+1)C_G} \eta_{t} \nabla_V \cL \left(f(V_{nC_G} ,k_{t} ),v_{t}\right)\;\;\; n=\{0,\ldots, L // C_G \} 
\end{align}

\paragraph{Local Memory Update.} The local memory W operates in parallel on sequence shards of length $S_L$. Within each shard, updates use a smaller chunk size $C_L$. 
{\small
\begin{align} \label{eq:tnt_local_memory_compress}
& W_t \gets \left\{
\begin{array}{ll}
     W_{\textrm{init}} & \textrm{if}\;\; 0 \equiv t \pmod{S_L} \\
     W_{t-1} - \sum_{\tau=\xi(t, C_{L})}^{t}   \eta_{\tau} \nabla_W \cL \left(f(W_{\xi(t, C_{L})} ,k_{\tau} ),v_{\tau}\right)& \textrm{Otherwise}  
\end{array} \right.
\end{align}
}%
\end{c2box}

The global memory update (Eq.~\ref{eq:tnt_global_memory_compress}) follows a standard chunkwise formulation where the state is carried over sequentially between large chunks. To maximize training throughput, the gradient for all tokens within a global chunk is computed with respect to the initial state of that chunk, allowing for a highly parallelized update.

In contrast, the local memory update (Eq.~\ref{eq:tnt_local_memory_compress}) introduces our key innovation: a periodic state reset. This rule enforces that the local memory state, $W_t$, is reset to a shared, learnable initial state $W_{init}$ at the beginning of each segment of length $S_L$. This periodic reset is the critical mechanism that breaks the long-range sequential dependency across the input, thereby enabling true context parallelism for the fine-grained local memory modules.

The hierarchical design of deep memory modules boosts training efficiency through a two-pronged approach. Global modules create hardware-saturating, compute-intensive operations by processing large chunks. Concurrently, the local memory's reset mechanism enables context parallelism, where the sequence is processed as independent chunks that can be distributed across devices or stacked on a single accelerator to substantially increase training throughput.

\subsubsection{TNT Memory Retrieval: Q-K Projection}\label{subsubsec:MemRetrieval}

As identified in Challenge 2, the memory compression step (Eq.~\ref{eq:tnt_local_memory_compress}) optimizes $f(W, \cdot)$  to map the key space to the value space. However, at retrieval, the network is queried using a query vector, $q_t$, which may lie outside the learned key domain, degrading performance.

To resolve this, we propose \textit{Q-K Projection}: projecting the query $q_t$ onto the subspace spanned by previously observed keys. This ensures the input to the memory function is in the space  memory was trained on. The final output combines retrieval from the global memory (raw query) and the local memory (projected query). We apply projection only locally as its fine-grained nature makes it more sensitive to the mismatch

\begin{c3box}
\noindent
\textbf{TNT Memory Retrieval Rule.} The hierarchical memory is retrieved as follows:
\begin{align}
    o_t = f(V_{\xi(t, C_G)}, q_t) + f\left(W_t, \sum_{\tau=\xi(t, C_{L})}^t \frac{k_\tau k_\tau^\top}{\|k_\tau\|^2} q_t\right) 
\end{align}

\end{c3box}
Crucially, this Q-K Projection does not require storing all past keys, which would be computationally and memory prohibitive. Instead, the projection matrix, $\sum_{\tau=1}^t \frac{k_\tau k_\tau^\top}{\|k_\tau\|^2} \in \mathbb{R}^{d\times d}$, can be maintained as a running sum. This results in a constant-size state that is updated efficiently in a chunkwise parallel manner. Since many modern deep memory modules normalize the query $\left(q_t\right)$ and key $\left(k_t\right)$ vectors by their L2 norm, the denominator in the Q-K projection can simplify to $\sum_{\tau=1}^t k_\tau k_\tau^{\top}$. We provide further details on this efficient implementation in Appendix~\ref{app:qk-implementation}.

\subsection{TNT Stage 2: Performance-focused Fine-tuning at Finer Resolution}\label{subsec:Stage2}

Having addressed training efficiency in Stage 1, we now turn to optimizing for inference performance. An intuitive approach to enhance model resolution would be to evaluate the pre-trained model using a smaller chunk size. However, as established in Challenge 3, a direct mismatch between the pre-training and evaluation chunk sizes leads to significant performance degradation.

Our key insight is that this train-test discrepancy can be rectified with minimal computational overhead. We empirically observe that a brief fine-tuning phase, where the pre-trained model is updated for a small number of steps using a smaller local memory chunk size, not only recovers but often surpasses the original performance.

Based on this finding, we introduce Stage 2 of our TNT framework: a \textbf{Performance-focused Fine-tuning Stage}. In this stage, we continue training the efficiently pre-trained model with a smaller local chunk size ($C_ L^{\prime} <C_{L}$). This process adapts the model to the higher resolution required for optimal inference at a fraction of the cost of pre-training. By doing so, Stage 2 directly resolves Challenge 3, bridging the gap between the large chunk sizes required for efficient training and the small chunk sizes that yield the best performance at inference.

This two-stage process decouples pre-training efficiency from inference requirements. The bulk of training uses maximum throughput (large chunks), while the final model is produced with minimal overhead. Furthermore, fine-tuning specializes the model for the ideal inference scenario: a local chunk size of one $(C_{L}^{\prime}=1)$. This aligns with the standard prefill-and-decode paradigm of autoregressive generation. The global memory handles the context prefill, and the optimized local memory handles iterative decoding.

\section{Experiments}\label{sec:experiments}
We empirically evaluate our two-stage training framework, TNT. While TNT is model-agnostic, we instantiate it with a strong deep memory model, Titans~\citep{behrouz2024titans}, to demonstrate its effectiveness. We validate claims about training time and model accuracy in our  experiments.

\subsection{Experimental Setup}

\textbf{Baselines.}
We compare against several strong long-context architectures. Our primary comparison is  Titans ~\citep{behrouz2024titans}, our base model. We also benchmark against vanilla Transformer ~\citep{vaswani2017attention}, Gated Transformer~\citep{qiu2025gated}, and TTT~\citep{sun2024learning}.

\textbf{Training and TNT Configuration.}
We train 150M parameter models following ~\citep{behrouz2024titans}, using a T5 tokenizer (32k vocab). We use the AdamW optimizer ~\citep{loshchilov2017decoupled} with 0.1 weight decay and a cosine schedule (peak LR $1\times10^{-3}$). Experiments are conducted on a TPUv5 pod (2x2x2 topology, model parallelism 2). For TNT, the $N$ local modules configuration is denoted by their chunksizes, $C_{L}=\{C_{L,1},...,C_{L,N}\}$. For instance, $C_L=\{8, 16\}$ indicates two local modules with chunksizes 8 and 16. The global memory uses $C_{G}=2048$.

\textit{Experimental Configurations.} For efficiency benchmarks (Sec.~\ref{subsec:training_speed}), we vary context length (2k-32k) with a 0.5M token batch size and local window $S_{L}=2048$. For performance evaluation (Sec~\ref{subsec:model_performance}), we use a 16k context length, 1M token batch size, and $S_{L}=4096$.

\begin{table*}[!t]
\centering

\caption{TNT reaches the target training loss up to 17× faster than the baseline Titans. The table compares the time required for different 150M models to reach the same target loss $3.20$.}
\begin{adjustbox}{max width=0.8\textwidth}
 \begin{tabular}{l l r | c c}
            \toprule
            \textbf{Models} & \textbf{Implementation}  &  $C$ or $C_L$  & \textbf{Training Time (hrs)} & \textbf{Speedup} \\
            
            Titans  & JAX & 8  & 19.48 & $1.00\times$ \\

            Titans  & JAX & 16 & 10.79 &  $1.81\times$ \\
            Titans  & JAX & 32  & 6.45 &  $3.02\times$ \\
            Titans  & JAX & 64 & 4.18 & $4.67\times$ \\
            Titans & JAX & 128  & 3.71 & $5.25\times$ \\
            Transformer (w/o gating)     & JAX & -  & 1.74 & $11.18\times$  \\
            Transformer (w gating) & JAX & -  & 1.38 & $14.10\times$ \\ 
            Transformer (w/o gating)        & FlashAttention (Pallas) & -   & 1.23 & $15.90\times$  \\
            Transformer (w gating) & FlashAttention (Pallas) & - & 0.96 & \colorbox{myblue}{$20.22\times$} \\ 
            \midrule
            TNT & JAX & \{8\}   & 2.54 & $7.68\times$ \\
            TNT & JAX & \{16\}   & 1.65 & $11.78\times$ \\
            TNT & JAX & \{32\}  & 1.22 & $ 15.92\times$ \\
            TNT & JAX & \{64\}  & 1.12 & \colorbox{myblue}{$ 17.37\times$} \\
            TNT & JAX & \{128\}  & 1.16 & $16.75\times$ \\
            \bottomrule 
        \end{tabular}
        \label{tab:speed-up}
\end{adjustbox}
\end{table*}

\subsection{Faster Memory Training with TNT} \label{subsec:training_speed}

\begin{wrapfigure}{r}{0.52\linewidth}
    \centering
    \includegraphics[width=\linewidth]{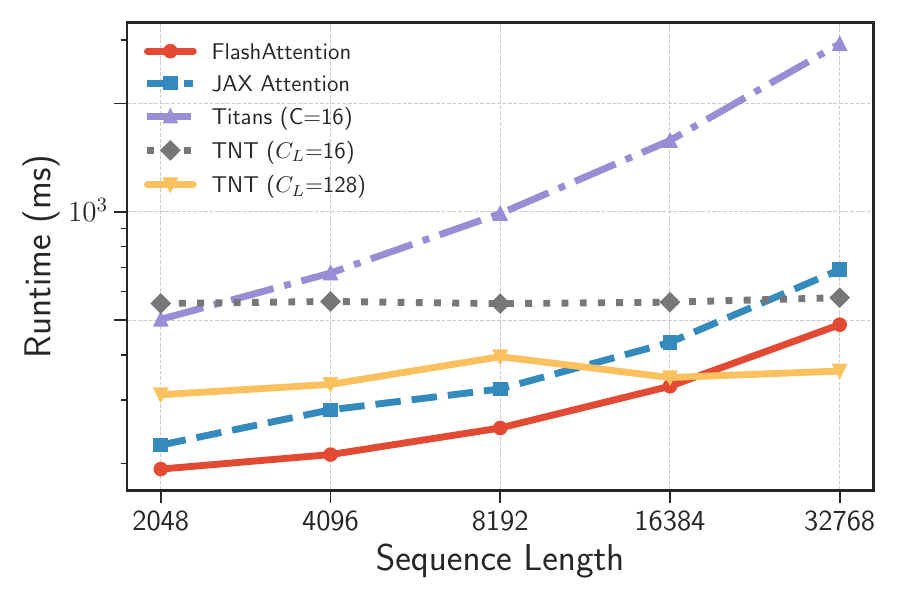}
    \caption{Runtime comparison of different models and implementations across varying sequence lengths, with the number of tokens per batch fixed at 0.5M. Additional results are presented in \autoref{fig:throughput}.}
    \label{fig:throughput_partial}
\end{wrapfigure}

\paragraph{Linear Runtime Scaling with Sequence Length.}
We first analyze single-step runtime performance by varying the sequence length while keeping the number of tokens per batch fixed. As shown in \autoref{fig:throughput_partial}, TNT's runtime grows linearly with sequence length, in contrast to the quadratic growth of Titans and standard attention. This scaling advantage is significant at long contexts. At a 32K sequence length, TNT is \textbf{5.1$\times$ faster} than a comparable Titans model with the same memory chunksize ($C_L=C=16$). We also observe that larger local chunk sizes consistently improve TNT's speed; with $C_L=\{128\}$, TNT is \textbf{1.3$\times$ faster} than the highly optimized FlashAttention kernel~\citep{dao2024flashattention}.

TNT's highly parallelizable architecture achieves a runtime that scales linearly with sequence length, a key advantage over the quadratic complexity of standard attention. Although models like Titans are also theoretically linear, their inherent sequential dependencies impede effective parallelization, resulting in poor hardware utilization and slower wall-clock times on long sequences. As sequence length increases, TNT's superior scalability creates a crossover point where it becomes significantly faster. This efficiency is most pronounced at very long contexts; for instance, at a sequence length of $32 K$, a native JAX implementation of TNT ($C_L=128$) outperforms even the highly optimized FlashAttention kernel, confirming its suitability for demanding long-context training scenarios.

\paragraph{Time-to-Quality Comparison.}
We next translate these single-step runtime gains into a practical time-to-quality setting. As shown in \autoref{tab:speed-up}, our TNT framework significantly accelerates the total training time required to reach a target model quality. Our best configuration achieves this up to \textbf{17.4$\times$ faster} than the original Titans baseline. This efficiency gain is fundamental to the architecture; for instance, using an identical local memory chunksize of 8, TNT is already \textbf{7.7$\times$ faster} than its Titans counterpart. While competitive with standard vanilla Transformers in JAX, our implementation does not yet outperform highly optimized baselines like the Gated Transformer with FlashAttention~\citep{dao2022flashattention}. This is an expected result, as TNT currently lacks a custom kernel, which we leave for future work. Nonetheless, these results establish TNT as an efficient foundation for research on recurrent models, with a clear path toward matching the speed of state-of-the-art Transformers.

\begin{table*}[!tb]
\centering

\caption{ 
Performance of TNT (150M parameters) and baselines on language modeling and common-sense reasoning tasks, trained on 10B tokens, respectively. For TNT models, the global chunksize is fixed at $C_G=2048$. $C_L$ denotes the set of chunksizes used by the local memory modules. The best results within a block are \colorbox{myblue}{highlighted}. The detailed training time is reported in Table~\ref{tab:tnt-training-time}}
\label{tab:tnt-main-results}
\begin{adjustbox}{max width=1\textwidth}
\begin{tabular}{l| r | c c c c | c c c c c }
\toprule
\textbf{Model}  & $C$ or $C_L$& \textbf{C4}  & \textbf{FineWeb} &  \textbf{PG19} & \textbf{Avg.} & \textbf{PIQA} &    \textbf{Hella.} &  \textbf{ARC-e} &  \textbf{CSQA} & \textbf{Avg.} \\
 & & ppl $\downarrow$  &  ppl $\downarrow$ &  ppl $\downarrow$  & ppl $\downarrow$ & acc $\uparrow$  & acc $\uparrow$  & acc $\uparrow$  & acc $\uparrow$  & acc $\uparrow$ \\
\midrule
\midrule
\multicolumn{9}{c}{\textbf{150M params / 10B tokens}} \\
\midrule
Transformer (w/o gating) & - & 20.98 & 20.59 & 29.18 & 23.58 & 62.0 & 30.9 & 34.8 & 25.5 & 38.3  \\
Transformer (w gating) & - & \colorbox{myblue}{19.82} & \colorbox{myblue}{19.61} & \colorbox{myblue}{27.75} & \colorbox{myblue}{22.39}  & \colorbox{myblue}{63.3} & \colorbox{myblue}{32.2} & \colorbox{myblue}{36.8}  & 26.7 & \colorbox{myblue}{39.7} \\
TTT (\citeyear{sun2024learning})  & 256 & 24.18 & 24.31 & 34.36 & 27.62  & 60.6 & 30.8 & 34.1 & 26.9 & 38.1  \\
Titans (\citeyear{behrouz2024titans}) \footnotemark  & 256 & 23.53 &  24.13 & 33.73 &  27.13 & 61.3 & 30.8 & 35.1 & \colorbox{myblue}{27.8} & 38.8 \\
Titans & 8 & 22.25 & 22.07  & 30.90 & 25.07 & 60.8  &32.0 &35.5  & 27.8 &39.0  \\
\midrule 
\multicolumn{9}{c}{\textbf{TNT Stage 1: Efficiency-Focused Pre-training} } \\
\midrule
TNT Stage 1  & \{8\} & 21.04 & 21.01 & 30.24 & 24.10  & 61.8 & 32.8 & 37.4 & 30.3 & 40.6 \\
      & \{8,16\} & 20.74	& 20.73	& 29.94& 23.80 & \colorbox{myblue}{63.5} & 32.4 & \colorbox{myblue}{37.4}& \colorbox{myblue}{30.6} & \colorbox{myblue}{41.0}   \\
     & \{4,8,16\} & 20.47 & 20.43 & 29.43 & 23.44 & 62.9 & \colorbox{myblue}{32.4} & 36.4 & 28.9 & 40.2  \\
    & \{4,8,16,32\} & \colorbox{myblue}{20.15} & \colorbox{myblue}{20.17} & \colorbox{myblue}{29.08} & \colorbox{myblue}{23.13} & 63.2 & 32.0 & 36.7 & 30.3 & 40.6 \\
\midrule
\multicolumn{9}{c}{\textbf{TNT Stage 2: Performance-Focused Fine-tuning on Stage 1 models} } \\
\midrule
TNT Stage 2  & \{1\} & 20.86 & 20.91 & 30.21 & 23.99 &  63.2 & \colorbox{myblue}{32.8} & 37.4 & 30.1 & 40.9 \\
             & \{2,4\} & 20.65 & 20.70 & 29.97 & 23.77 & 63.4 & 32.5 & 37.3 & 30.2 & 40.9 \\
            & \{2,4,8\} & 20.32  & 20.35 & 29.42 & 23.36 & \colorbox{myblue}{64.0} & 32.0 & 36.9 & 28.1 & 40.3\\
             & \{2,4,8,16\} & \colorbox{myblue}{20.10} & \colorbox{myblue}{20.13} & \colorbox{myblue}{29.05} & \colorbox{myblue}{23.09} & 63.5 & 32.3 & \colorbox{myblue}{37.4} & \colorbox{myblue}{30.2} & \colorbox{myblue}{40.9} \\

 \bottomrule
\end{tabular}
\end{adjustbox}
\label{tab:partial_main-table-glam}
\end{table*}
\afterpage{\footnotetext{This work is based on our re-implementation of the TTT/Titans models. The models are trained in the same setup for a fair comparison.}}

\vspace{-0.5cm}
\subsection{TNT Improves Model Quality}\label{subsec:model_performance}

Our TNT framework significantly enhances model quality, outperforming strong RNN-based baselines and standard Transformer implementations. As detailed in \autoref{tab:tnt-main-results}, the initial \textbf{Stage 1 pre-training} is highly effective on its own. Our best Stage 1 model achieves an average perplexity of \textbf{23.13}, a marked improvement over the best-performing Titans model (25.07) and the vanilla Transformer (23.58). While TNT does not fully match the perplexity of the state-of-the-art Gated Transformer (22.39), it achieves a higher average accuracy on common-sense reasoning tasks (\textbf{41.0\%} vs. 39.7\%). At this scale, we consider perplexity a more stable metric for language modeling capability, as downstream task accuracy can be subject to higher variance.

Furthermore, the \textbf{Stage 2 fine-tuning} process offers an efficient method to further boost performance. This stage is computationally inexpensive, requiring only an additional 5\% of the original pre-training compute (see \autoref{tab:tnt-training-time}), yet it consistently lowers the average perplexity to a final value of \textbf{23.09}. These results validate TNT as an effective framework for producing high-quality models that surpass the limitations of prior RNN-based architectures and stand as a strong alternative to standard Transformers.

\subsection{Ablation Study} \label{sec: ablation_study}

\begin{wraptable}{r}{0.5\textwidth}
    \centering
    \vspace{0cm}
    \caption{Ablation study on TNT. The results show the contribution of each proposed change to the deep memory modules.
    }
    \label{tab:ablation-results}
    \begin{adjustbox}{max width=0.5\textwidth}
        \begin{tabular}{l c c c}
        \toprule
        \textbf{TNT} & $N$  & \textbf{Language Modeling} & \textbf{C.S. Reasoning} \\
        & & ppl $\downarrow$ & acc $\uparrow$ \\
        \midrule
        Base Model (Titans)           & -    & 23.53 & 38.8  \\
        \midrule
        \midrule
        \multicolumn{4}{l}{\textbf{TNT Stage 1 (\texttt{1 Global Memory})}} \\
        \hspace{4pt}+\texttt{1 Local Memory} & 1 & 21.04 & 40.6 \\
        \hspace{4pt}+\texttt{2 Local Memory} & 2 & 20.74 & 41.0 \\
        \hspace{4pt}+\texttt{3 Local Memory} & 3  & 20.47 & 40.2 \\
        \hspace{4pt}+\texttt{4 Local Memory} & 4  & 20.15 & 40.6 \\
        \midrule
        \midrule
        TNT Stage 1 & 1 & 21.04 & 40.6 \\
        w/o global memory & - & 25.60 & 35.5 \\
        w/o Q-K projection & 1 & 22.01 & 36.4\\
        w Stage 2  & 1 & 20.86 & 40.9 \\
        \bottomrule
    \end{tabular}
    \end{adjustbox}
\end{wraptable}

We conducted an ablation study to validate TNT's key design choices, with results summarized in Table~\ref{tab:ablation-results}. 

\textbf{Hierarchical Memory.} The effectiveness of our hierarchical design is evident. Incrementally adding local memory modules consistently improves performance over the Titans baseline, reducing perplexity from 23.53 to 20.15 with four local modules. Conversely, removing the global memory is detrimental (PPL increases to 25.60), confirming its critical role in capturing long-range dependencies that are otherwise lost due to the local memories' reset mechanism.

\textbf{Q-K Projection.} The query-key projection proves essential for performance. Its removal incurs a substantial penalty (PPL increases from 21.04 to 22.01), validating our hypothesis that it is necessary to mitigate the compression-retrieval mismatch (Challenge 2). 

\textbf{Stage 2 Fine-tuning.} Applying Stage 2 fine-tuning further enhances model capabilities, improving both language modeling (20.86 PPL) and common-sense reasoning ($40.9$\%). This demonstrates its effectiveness in adapting the pre-trained models for high-resolution inference.

\section{Conclusion} \label{sec: conclusion}

We introduce TNT, a two-stage training framework that resolves the fundamental conflict between training efficiency and inference performance in deep memory modules. By leveraging a hierarchical memory architecture with periodic state resets, TNT enables massive context parallelism during pre-training, followed by efficient fine-tuning for high-resolution inference. Our experiments demonstrate up to a 17$\times$ speedup compared to the most accurate RNN baselines while simultaneously improving performance. TNT removes a critical scalability bottleneck, significantly improving the practicality of deep memory modules and facilitating future research to close the performance gap with Transformers.

\printbibliography
\newpage
\appendix
\section{Related Work}\label{app:rw}

\textbf{Modern Linear Recurrent Neural Networks}
Due to the quadratic complexity of transformers, recently developing alternative architectures have gained attention, which led to the development of efficient recurrent alternatives~\citep{tiezzi2024resurgence}. Initial advancements in this domain, starts with models such as RetNet~\citep{sun2023retentive}, RWKV~\citep{peng2023rwkv}, and S5~\citep{smith2023simplified}, which employed data-independent transition matrices coupled with Hebbian-like update mechanisms. Subsequently, a second generation of models emerged, incorporating input-dependent parameters within these linear architectures (e.g., linear RNNs~\citep{hasani2023liquid, smith2023simplified}, RWKV6~\citep{peng2024eagle}). These models also explored more expressive memory updating rules, notably those based on the delta rule~\citep{peng2025rwkv7, schlag2021linear, yang2024parallelizing, yang2024gated, liu2024longhorn}. Further evolution in this line of research has extended these memory architectures to deeper models, while concurrently utilizing delta-rule-like update mechanisms~\citep{sun2024learning} or data-dependent momentum-based update rules with forget gating~\citep{behrouz2024titans}. More recently, to augment the performance of delta-rule-based sequential models, \citet{siems2025deltaproduct} have proposed the application of multiple gradient descent updates per token, thereby yielding more expressive sequence models, particularly in state tracking tasks. In addition to the above fast linear recurrent sequence models, several studies have focused on RNNs with non-linear recurrence~\citep{behrouz2024titans,behrouz2025itsconnectedjourneytesttime,  behrouz2025atlas, csordas2024recurrent, merrill2024the, lim2024parallelizing,  schone2025implicit, karami2025lattice, gonzalez2024towards}, and how their training can be faster~\citep{gonzalez2024towards, lim2024parallelizing, schone2025implicit}.

\textbf{Fast Weight Programs}
The conceptualization of linear layers as key-value associative memory systems can be traced back to Hopfield networks~\citep{hopfield1982neural}. This concept was subsequently developed in the context of fast weight programmers, wherein dynamic fast programs are integrated into recurrent neural networks to serve as writable memory stores~\citep{schlag2021linear, schmidhuber1992learning, schmidhuber1993reducing}. Among the learning paradigms for such systems, Hebbian learning~\citep{hebb2005organization} and the delta rule~\citep{prados1989neural} have emerged as the most prominent. Both learning rules have been the subject of extensive investigation within the existing literature~\citep{munkhdalai2017neural, schmidhuber1992learning, munkhdalai2019metalearned, schlag2021linear, irie2021going, yang2024parallelizing, yang2024gated}.

\textbf{Hopfield Networks} 
Our formulation is architecturally founded upon the broad concept of associative memory, wherein the primary objective is to learn an underlying mapping between keys and values. Seminal work by \citet{hopfield1982neural} on Hopfield Networks introduced one of the earliest neural architectures explicitly based on associative memory, defining it through the minimization of an energy function for storing key-value pairs. Although traditional Hopfield networks have seen diminished applicability in recent years, primarily due to constraints in vector-valued memory capacity and the nature of their energy function, several contemporary studies have focused on enhancing their capacity through various methodologies. These include efforts by \citet{krotov2021hierarchical}, \citet{li2024expressive}, and \citet{krotov2016dense}. Notably, extensions to the energy function of these models, often incorporating exponential kernels, have been explored~\citep{krotov2016dense, lucibello2024exponential}. Furthermore, the relationship between these modernized Hopfield networks and Transformer architectures has been a subject of recent investigation~\citep{ramsauer2021hopfield, hu2024provably}.

\section{Connection of QK-Projection with Memory Bounded Transformers} \label{app:qk-connection}
Revisiting our QK-Projection retrieval process, we first project the query vector into the space of stored keys and then retrieve its corresponding value by a forward pass on the memory module. In particular, given query $q_t$ and stored keys of $\{ k_i\}_{i = 1}^{t}$, the output corresponds to query $q_t$ can be calcualted as:
\begin{align}
    o_t = f\left(W_t, \sum_{\tau=1}^t \frac{k_\tau k_\tau^\top}{\|k_\tau\|^2} q_t\right). 
\end{align}
Given a normalization of keys, i.e., $\|k_{\tau}\|_2 = 1$, this formulation, can be re-written as:
\begin{align}
    o_t = f\left(W_t, \sum_{\tau=1}^t k_\tau k_\tau^\top q_t\right),
\end{align}
in which the second element, $\sum_{\tau=1}^t k_\tau k_\tau^\top q_t$, is equivalent to a simple forward pass for query $q_t$ over a linear memory module of $\mathcal{M}'_{t} = \sum_{\tau=1}^t k_\tau k_\tau^\top$ with recurrence of:
\begin{align}
    \mathcal{M}'_{t} = \mathcal{M}'_{t-1} + k_t k_t^\top.
\end{align}
Such formulation of QK-Projection can remind us the two-pass process of memory bounded Transformers~\citep{peng-etal-2022-abc, zhang2024gated, karami2025trellis}, where in the simple linear attention form~\citep{peng-etal-2022-abc}, the retrieval process can be written as:
\begin{align}
    &W_t = W_{t-1} + \varphi_t v_t^{\top} , \\
    &o_t = W_t  \: \texttt{softmax} \left(\left( \sum_{\tau = 1}^{t} k_{\varphi} \varphi_{\tau}^{\top}   \right) q_t \right). 
\end{align}
Comparing with above two-pass process of ABC~\citep{peng-etal-2022-abc},  our QK-projection method is applicable to both deep and linear memory. Furthermore, parameters of $\varphi_t$ as well as $k_t$ are tied and so the learning process is considerably easier, making the model more adaptable to new data/tasks. Moreover, when employ this projection in the local memory, we only do the summations starting from the ``reset" state of the memory rather than starting from $\tau=1$.

\section{Efficient Implementation of QK-Projection} \label{app:qk-implementation}
This section details the efficient, parallelizable implementation of the QK-Projection mechanism. We demonstrate that this projection can be integrated into the chunkwise training paradigm without introducing sequential bottlenecks, thereby preserving the training efficiency of the TNT architecture.

The QK-Projection relies on a projection matrix, $\mathcal{M}_t$, which accumulates the outer products of keys within the current local memory shard (length $S_L$). Assuming normalized keys ($\|k_\tau\|=1$), this matrix is defined by the following recurrence:
$$
\mathcal{M}_t = 
\begin{cases}
    k_t k_t^\top & \text{if } t \equiv 1 \pmod{S_L} \\
    \mathcal{M}_{t-1} + k_t k_t^\top & \text{otherwise}
\end{cases}
$$
This rule ensures that the projection state $\mathcal{M}_t$ is reset at the beginning of each shard, mirroring the reset of the local memory state $W_t$. The local memory retrieval is then computed as $f(W_t, \mathcal{M}_t q_t)$.

\paragraph{Chunkwise Parallel Computation.}
To maintain training efficiency, $\mathcal{M}_t$ is computed in a chunk-parallel manner. For any time step $t$ within a chunk of size $C_L$, the projection matrix can be decomposed into two components:
$$
\mathcal{M}_t = \underbrace{\mathcal{M}_{\xi(t, C_L)-1}}_{\text{Carry-over State}} + \underbrace{\sum_{\tau=\xi(t, C_L)}^t k_\tau k_\tau^\top}_{\text{Intra-chunk Sum}}
$$
The first term is the final state from the previous chunk, which is carried over. The second term, the intra-chunk sum, is computed efficiently for all steps in the chunk simultaneously using a parallel prefix sum (scan) operation over the sequence of outer products $\{k_\tau k_\tau^\top\}$.

This implementation preserves end-to-end parallelism. The state passed between chunks is a single, constant-size matrix ($d \times d$), incurring minimal overhead. The periodic reset is handled by re-initializing this carry-over state at shard boundaries. Thus, QK-Projection enhances the model's retrieval mechanism without compromising the training efficiency fundamental to the TNT architecture.

\section{TNT Applicability} 
In this paper, we have focused on showing the effectiveness and efficiency of TNT and so for the sake of clarity, we use a simple memory module that optimizes its inner-loop with gradient descent. However, TNT recipes are applicable to different deep memory and non-linear architectures. For example, one can adapt the gating formulation in Titans~\citep{behrouz2024titans} or Mamba2~\citep{dao2024transformers} for each of the local memories as well as the global memory. Another potential exploration is to incorporate closed feedback loop in the objective of the inner-loop as it has done in \citet{hu2025comba}. Similarly, one can employ more expressive optimizers as the inner-loop optimizers such as gradient descent with momentum, AdamW~\citep{kingma2014adam}, or muon~\citep{jordan2024muon} as it has been done by \citet{behrouz2025atlas, zhang2025test}. While exploring all these combinations with TNT is a promising direction, for the sake of clarity and space constraint, we leave them for future studies.

\section{TNT Generalization Formulation} \label{app:tnt_complete}

The TNT architecture can be generalized to a hierarchical system comprising one global memory and $N$ parallel local memories. This allows the model to capture information at multiple timescales and resolutions simultaneously. Each local memory, denoted by $W^{(i)}$ for $i \in \{1, \ldots, N\}$, operates with its own distinct chunk size $C_{L_i}$, shard length $S_{L_i}$, and learnable initial state $W^{(i)}_{\text{init}}$.

\subsection{Generalized Memory Update}
The update rules are extended as follows: the single global memory evolves sequentially, while the $N$ local memories are updated in parallel, each with its independent schedule.

\paragraph{Global Memory Update.} The global memory state $V$ is updated sequentially with a large chunk size $C_G$, identical to the base case.
\begin{align} \label{eq:tnt_global_memory_general}
V_{(k+1)C_G} \gets V_{kC_G} - \sum_{t=kC_G}^{(k+1)C_G} \eta_{t} \nabla_V \mathcal{L} \left(f(V_{kC_G}, k_{t}), v_{t}\right)
\end{align}

\paragraph{N-Local Memories Update.} Each of the $N$ local memories $W^{(i)}$ operates in parallel. The state of each memory is reset periodically according to its specific shard length $S_{L_i}$, enabling multi-resolution context parallelism.
\begin{align} \label{eq:tnt_local_memory_general}
W_t^{(i)} \gets \left\{
\begin{array}{ll}
    W^{(i)}_{\textrm{init}} & \textrm{if } 0 \equiv t \pmod{S_{L_i}} \\
    W_{t-1}^{(i)} - \sum_{\tau=\xi(t, C_{L_i})}^{t} \eta_{\tau} \nabla_{W^{(i)}} \mathcal{L} \left(f(W^{(i)}_{\xi(t, C_{L_i})}, k_{\tau}), v_{\tau}\right) & \textrm{Otherwise}
\end{array} \right.
\end{align}
where $i = 1, \ldots, N$.

\subsection{Generalized Memory Retrieval}

During retrieval, the final output is a composition of the outputs from the global memory and all $N$ local memories. The global memory uses the raw query $q_t$, while each local memory uses a Q-K projection tailored to its specific context window, determined by its chunk size $C_{L_i}$.

\paragraph{TNT Generalized Retrieval Rule.} The hierarchical memory is retrieved by summing the contributions from each memory module.
\begin{align} \label{eq:tnt_retrieval_general}
o_t = f(V_{\xi(t, C_G)}, q_t) + \sum_{i=1}^{N} f\left(W_t^{(i)}, \sum_{\tau=\xi(t, C_{L_i})}^t \frac{k_\tau k_\tau^\top}{\|k_\tau\|^2} q_t\right)
\end{align}
This formulation allows the network to integrate long-range dependencies from the global memory with fine-grained, parallel-processed information from a diverse set of local memories, each specializing in different temporal patterns.

\begin{figure}[!t]
        \centering
        \includegraphics[width=1\linewidth]{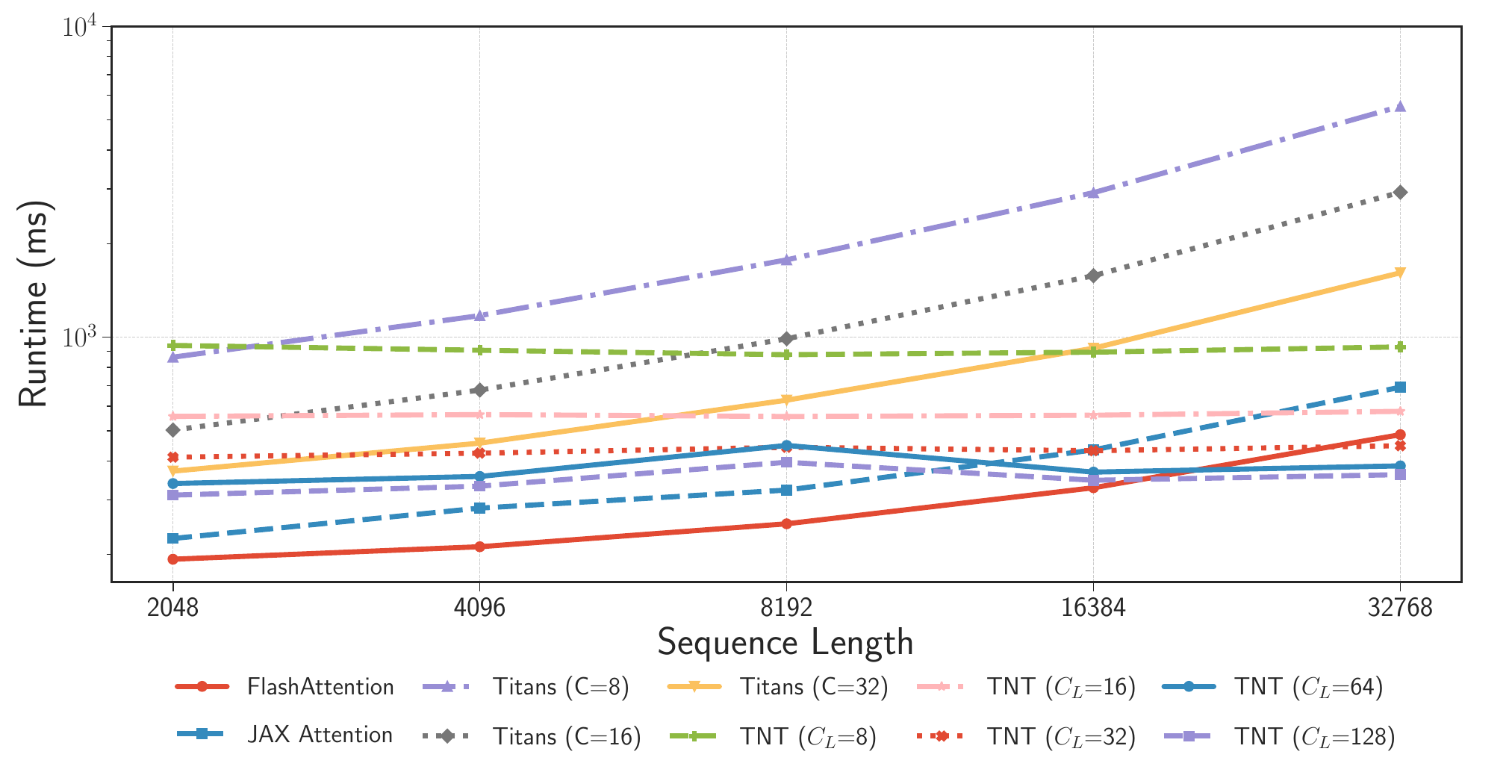}
        \caption{Runtime comparison of different models and implementations across varying sequence lengths, with the number of tokens per batch fixed at 0.5M.}
        \label{fig:throughput}
\end{figure}

\begin{table*}[!tb]
\centering
\caption{Training time for 150M parameter models trained on 10B tokens. For TNT models, the global chunksize is fixed at $C_G=2048$, and $C_L$ denotes the set of chunksizes for the local memory modules.}
\label{tab:tnt-training-time}
\begin{adjustbox}{max width=0.5\textwidth}
\begin{tabular}{l l c}
\toprule
\textbf{Model} & $C$ or $C_L$ & \textbf{Training Time (hrs)} \\
\midrule
\midrule
\multicolumn{3}{c}{\textbf{150M params / 10B tokens}} \\
\midrule
Transformer (w/o gating) & - & 0.80 \\
Transformer (w gating) & - & 0.82 \\
TTT (\citeyear{sun2024learning}) & 256 & 1.69 \\
Titans (\citeyear{behrouz2024titans}) & 256 & 1.99 \\
Titans & 8 & 8.44 \\
\midrule
\multicolumn{3}{c}{\textbf{TNT Stage 1}} \\
\midrule
TNT Stage 1 & \{8\} & 3.06 \\
& \{8,16\} & 4.24 \\
& \{4,8,16\} & 5.00 \\
& \{4,8,16,32\} & 5.55 \\
\midrule
\multicolumn{3}{c}{\textbf{TNT Stage 2}} \\
\midrule
TNT Stage 2 & \{1\} & 0.15 \\
& \{2,4\} & 0.23 \\
& \{2,4,8\} & 0.26\\
& \{2,4,8,16\} & 0.46\\
\bottomrule
\end{tabular}
\end{adjustbox}
\end{table*}

\end{document}